\documentclass{article}

\usepackage{arxiv}

\usepackage[utf8]{inputenc} 
\usepackage[T1]{fontenc}    
\usepackage{hyperref}       
\usepackage{url}            
\usepackage{booktabs}       
\usepackage{amsfonts}       
\usepackage{nicefrac}       
\usepackage{microtype}      
\usepackage{graphicx}
\graphicspath{ {./images/} }
\usepackage{amssymb}
\usepackage{pifont}
\newcommand{\cmark}{\ding{51}}%
\newcommand{\xmark}{\ding{55}}%
\DeclareUnicodeCharacter{FB01}{fi}
\DeclareUnicodeCharacter{FB02}{fi}

\title{A Compact Deep Architecture for Real-time Saliency Prediction}

\author{
 Samad Zabihi \\
  School of Electrical and Computer Engineering\\
  Shiraz University\\
  Shiraz, Iran \\
  \texttt{s.zabihi@shirazu.ac.ir} \\
   \And
 Hamed Rezazadegan Tavakoli\\
  Nokia Technologie\\
  Espoo, Finland \\
  \texttt{hamed.rezazadegan\_tavakoli@nokia.com} \\
  \And
 Ali Borji \\
  HCL America\\
  NYC, USA \\
  \texttt{aliborji@gmail.com} \\
}

\begin{document}
\maketitle
\begin{abstract}
Saliency computation models aim to imitate the attention mechanism in the human visual system. The application of deep neural networks for saliency prediction has led to a drastic improvement over the last few years. However, deep models have a high number of parameters which makes them less suitable for real-time applications. Here we propose a compact yet fast model for real-time saliency prediction. Our proposed model consists of a modified U-net architecture, a novel fully connected layer, and central difference convolutional layers. The modified U-Net architecture promotes compactness and efficiency. The novel fully-connected layer facilitates the implicit capturing of the location-dependent information. Using the central difference convolutional layers at different scales enables capturing more robust and biologically motivated features. We compare our model with state of the art saliency models using traditional saliency scores as well as our newly devised scheme. Experimental results over four challenging saliency benchmark datasets demonstrate the effectiveness of our approach in striking a balance between accuracy and speed. Our model can be run in real-time which makes it appealing for edge devices and video processing.

\end{abstract}


\section{Introduction}
The human brain has limited visual processing resources. To overcome this limitation, it has developed a mechanism to rapidly assign more resource to the more important subsets of the scene \cite{RN25}. The human visual system detects the salient objects of the scene and focuses on these salient subsets while ignoring the less salient parts \cite{RN23}. This happens in the brain in a fraction of a second. Saliency prediction models are instrumental to study mechanisms of visual attention, and also predict where people focus when they look at images or watch videos \cite{RN25}. The gaze prediction models have been used in many visual tasks such as object recognition \cite{p3}, image/video retargeting \cite{p4}, segmentation \cite{p5}\cite{p6}, visual tracking \cite{p7}, and image compression \cite{p8} across domains such as advertising, robotics, auto-driving, defense, game, assistive systems, and human-computer interaction. Early saliency prediction models are biologically-motivated and use low-level features \cite{RN29}\cite{itti2001computational}\cite{rw04} such as color, intensity, orientation, and texture. 
These techniques are unable to generally incorporate high-level features (e.g., contextual information, center prior and complex objects) and inherent correlation of various visual subsets in a scene (e.g., correlation of eyes, nose, ears, and mouth). The recent wave of saliency models are based on deep neural networks. The significance of deep learning has been such that some have categorized the methods to pre- and post deep learning \cite{RN25}. The deep learning-based methods have reduced the gap between model prediction and ground-truth significantly \cite{mitb}. On the one hand, saliency has been assumed a fast and computationally inexpensive process. On the other hand, deep models have a large number of parameters and require tremendous computational resources. SAM-ResNet \cite{RN17} for example, is a very huge deep model with about 70 million parameters and use recursive components. The unnecessary complexity of recent deep saliency models results in large inference time that is not preferable. Nevertheless, to our knowledge, there exists no evaluation of resource requirement of deep models for predicting saliency. These cases show the importance of studying saliency models in terms of complexity and introducing a new class of more compact and efficient models.\\

The recent progress in hardware capabilities of smart devices has made image and video processing on these platforms possible. However, the limitation of hardware resources still challenges the real-time processing of images and videos on these platforms, especially on CPU processors. The gaze prediction models can also be used to solve the resource assignment for image and video processing in real-time. Therefore, saliency models should be able to do the video and image processing in real-time. Resource requirement is an essential aspect for saliency computation, especially when hardware implementations are involved. Most of the state-of-the-art saliency models are very slow while many real-world applications require fast and highly efficient predictions. Given high resource requirement of deep models and the resource efficiency of pre-deep learning models, non-deep models are often the choice in hardware implementations\cite{Molin2020}.In this paper, (1) we study the computational requirement aspect for some of the most prominent state-of-the-art saliency prediction models; (2) To address the needs of real-time applications, we propose a new deep fast gaze prediction model. Our proposed model is highly efficient and is suitable for real-time applications. Our experimental results show that our model is very agile, even on ordinary CPUs. 

The remainder of the paper is organized as follows. The next section discusses related saliency prediction models. Section 3 presents our proposed saliency model. Section 4 describes some popular evaluation metrics, saliency datasets, and provides our implementation details. Section 5 presents ablation analysis of our model and reports the evaluation results of our proposed model over several saliency benchmark datasets. Finally, in section 6 we conclude.\\

\section{Related Works}
Visual attention has been studied for years. Numerous techniques have been proposed over the years to computationally model the human attention mechanism, e.g., using saliency models \cite{Tsotsos2011}. A saliency model predicts a density map that defines the location of conspicuous information in the input image/video. The ground truth density maps are often obtained by pooling together the information from fixations of several observers that are viewing the images (mostly under free-viewing condition). Thus, sometimes the task of predicting observers fixation density maps, a.k.a saliency prediction and saliency modeling, is referred to as fixation density prediction. 
This task is related to, but yet distinct from, saliency detection which aims to segment the most salient item in the scene \cite{Li2014} and shall not be confused.

\subsection{Classic saliency models}
Early saliency computation models were mostly inspired by human psychological and psychological properties and were based on “the feature integration theory” \cite{rw01}. Koch and Ullman \cite{rw02} is one of the first to build upon the feature integration theory. Itti and Baldi \cite{rw03} introduced a model based on Bayesian approaches. Harel et al. \cite{rw04} calculated saliency using graph theory. Hou and Zhang \cite{hou2007saliency} introduced a model based on frequency analysis. Some methods adopted an information-theoretic justification for attentive selection \cite{rw06}\cite{rw07}\cite{rw08}. Some other classic saliency models used machine learning algorithms \cite{rw10}\cite{rw11}\cite{RN14}.\\

\subsection{Deep saliency models}
Using deep neural networks (DNN) for saliency computation has made some drastic improvements \cite{RN25}. Vig et al.'s work \cite{RN28} was one of the first saliency models that used DNNs. In \cite{RN9} and \cite{RN10}, Kümmerer et al. used AlexNet and VGG-19 for extracting features from the input image. Kruthiventi et al. \cite{RN6} also used CNNs. Huang et al. \cite{RN15} and Cornia et al. \cite{RN16} integrated information at different image scales. Liu and Han \cite{RN18} and Cornia et al. \cite{RN17} used convolutional long short-term model (CNN-LSTM) to capture the global context of the scene. One of the most important drawbacks of these deep saliency models is that they have a large number of parameters and are computationally expensive for real-time applications. Note withstanding, we will discuss (in Section˜\ref{sec:framework}) some of the most popular methods that are re-implemented in our evaluation framework and has publicly available code.

\subsection{Non-deep fast saliency models}
In 2008, Butko et al. \cite{butko2008visual} simplified Zhang et al.’s Bayesian model \cite{zhang2007bayesian} to build a faster saliency model for real-time controlling of robot cameras.
Cui et al. \cite{cui2009temporal} used the idea of Spectral Residual \cite{hou2007saliency} to develop a fast motion saliency detection for video content analysis.
In 2011, Ho-phuoc et al. \cite{ho2011compact} proposed a compact model that uses the weighted combination of inter-frame difference, spatial contrast and central fixation bias to compute video saliency. 
Tavakoli et al. \cite{tavakoli2011fast} used sparse sampling and kernel density estimation for fast saliency computation of local feature contrast in a Bayesian framework. In \cite{ho2012compact}, Ho-phuoc et al. introduced another compact model for video-rate computation.
Wang et al. \cite{wang2013ship} proposed a saliency model based on random-forest for fast ship detection in high-resolution SAR images.
Li et al. \cite{li2017fast} introduced a fast video saliency detection model that predicts human fixations in compressed domain based on Residual DCT Coefficients Norm (RDCN) and Operational Block Description Length (OBDL). 
Qi et al. \cite{qi2016fast} proposed a fast saliency model based on gradient enhancement operation combined with Gaussian smoothing for real-time small target detection.

\subsection{Compact deep saliency models}
Designing fast and small deep neural models or obtaining such models from bigger neural networks is an ongoing research subject matter with extreme interest. 
It is argued that under constrained total parameter size, slim and deep neural networks usually outperform wide and shallow neural networks \cite{He_2016_CVPR}. 
 Several approaches have been proposed for obtaining a slim network from a higher capacity deep model, including, network pruning \cite{pruning}, knowledge distillation \cite{hinton2015distilling}, tensor decomposition \cite{lebedev2014speedingup}, etc. 
 
In deep saliency prediction, Theis et al. \cite{pruning} proposed employing pruning techniques to obtain a faster saliency model. To this end, they employ fisher pruning over DeepGazeII \cite{RN10} and remove the neurons until achieving a target network size and computational cost of 10.7 G-Flops (almost 10 times of our proposed model). After removing the neurons, the model is fine-tuned to recover from the performance loss. 

\subsection{Compact deep saliency detection}
In contrast to saliency prediction, where there has not been much attention to compact models, salient object detection community has investigated the compact models more often because of the application of segmentation task in mobile devices. Nevertheless, to our knowledge, those efforts have been mostly focused on proposing a neural architecture rather than understanding and analyzing the resource requirements. Note withstanding, their proposed models are for a distinct application and their methods may not directly apply to saliency prediction. We, however, provide a brief summary for the sake of completeness.
Xi et al. \cite{xi2017fast} proposed a compact saliency detection model based on a special deep convolutional neural network.  They constructed their deep model by modifying the VGG16-Net \cite{simonyan2014very}.  
Dabkowski et al. \cite{dabkowski2017real} developed a method for real time image saliency detection. Their method uses a black box classifier to learn which parts of the input image are deemed salient. They adopt a ResNet-50 under the U-Net architecture to extract feature maps from multiple resolutions and a masking model to manipulate the
scores of the classifier. Despite advances in the development of compact models, these models still suffer from their large number of parameters. 

\section{The Proposed Model}
\begin{figure}[hb!]
    \centering
    \includegraphics[width=.8\textwidth]{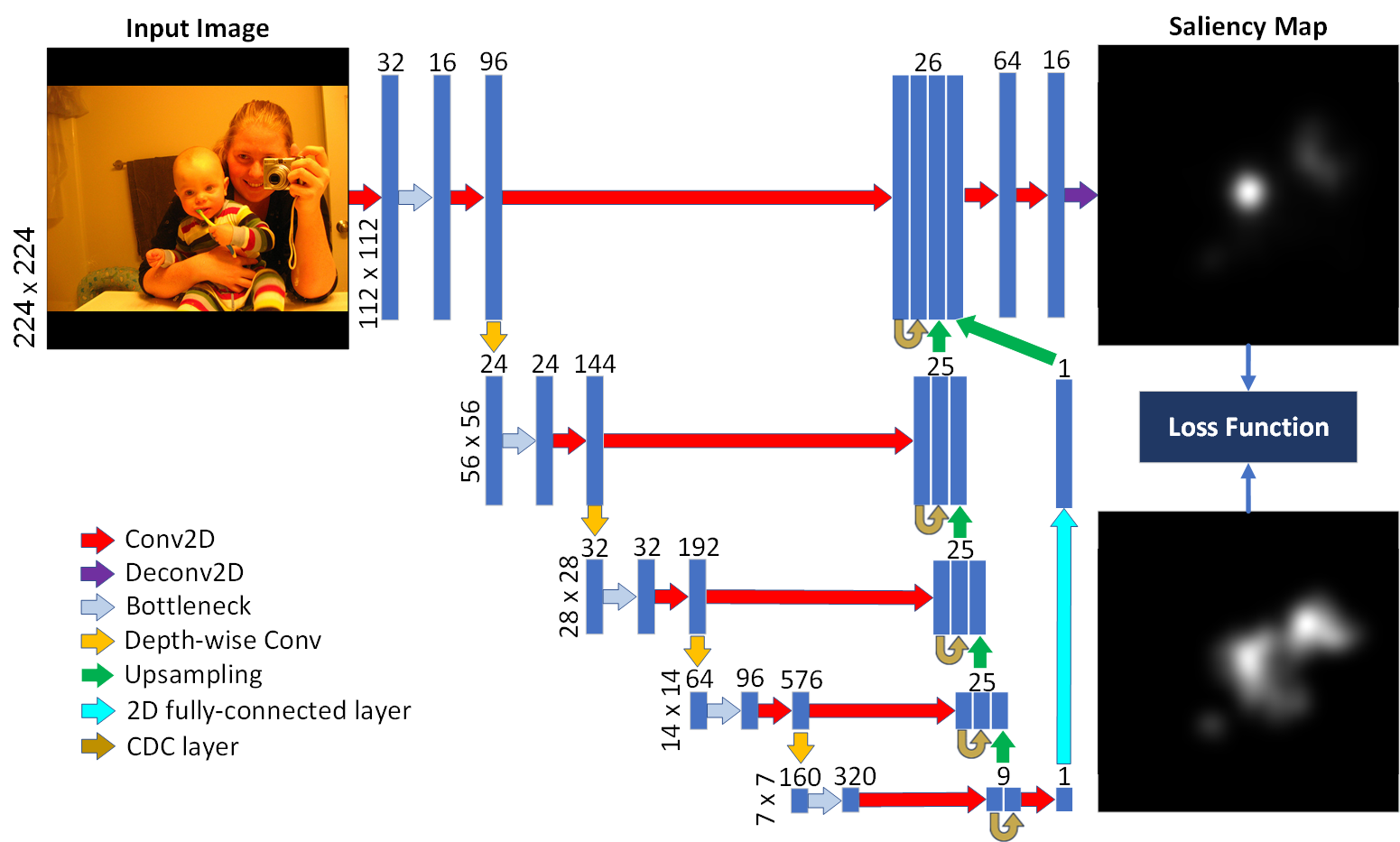}
    \caption{Our architecture for fast saliency computation. It is a asymmetric U-Net alike network that is composed of MobileNetV2 as backend, central difference Convolutional layers for extracting more consistent patterns and a fully-connected layer for extracting location-dependent features.}
    \label{fig:fig_c}
\end{figure}

Here we present a lightweight and computationally efficient architecture for fast saliency prediction. Our architecture reduces the number of parameters 13x compare to some state of the art models. Figure \ref{fig:fig_c} depicts our proposed architecture. Our architecture is a novel asymmetric U-Net alike network. It combines both fully-connected and convolutional layers. This modifications enables the model to capture global scene information, location dependent features and center prior. To promote efficiency and to keep the number of parameters small, we use depth-wise convolutions to reduce the number of parameters, resulting in an asymmetric U-Net like architecture.\\

The original U-Net architecture is an end-to-end fully convolutional network (FCN) \cite{RN3}. In this architecture, the feature maps are collected and concatenated from every resolution level. Most of this feature maps have a large number of feature channels and all of these channels are being collected in the upsampling part \cite{RN3} because of using copy and crop operators. This causes the final model to have many parameters, which is incompatible with our goal of creating a compact model. To fix some of these issues, we introduce an asymmetric U-Net like architecture which is customised for saliency prediction. In our architecture, we embedded a 2D fully-connected layer in the U-Net to compensate for some deficiencies in FCNs. This modification enables the model to capture global scene information, location-dependent features, and center prior. To keep the number of parameters limited, we used some extra convolutional layers that reduce the number of channels at each level. Our experimental result suggests that this method not only decreases the number of model parameters but also drastically reduces the FLOPS needed for saliency map prediction. It results in smaller inference time for the proposed model. To extract more consistent and biological motivated features from the scene and to increase feature diversity, we also employ central difference convolution (CDC) to capture intrinsic patterns of the scene.\\

Our proposed model consists of four components: MobileNetV2, U-Net architecture, central difference convolutional layers, and 2D fully-connected layer. Our model is fed by an RGB image with a size of 240x240 pixels. The model is composed of a contraction path and an expansion path that reduce and increase the resolution of feature maps respectively. In the contraction path, several feature maps are being extracted from the scene at diff rent resolution. In the expansion path, there is a central difference Convolutional layer at each step that extracts CDC features. A fully-connected layer is employed at the lowest resolution to capture some location-dependent information such as the center prior. The feature maps at different resolutions are up-sampled and concatenated, then some convolutional layers combine these feature maps to compute the final saliency map. At the training phase, the predicted saliency map is compared with the ground-truth fixation map by using a predefined loss function to compute the loss value. Then, the output error is back-propagated to train the network weights.

\subsection{MobileNetV2}
\begin{figure}[ht!]
    \centering
    \includegraphics[width=0.4\textwidth]{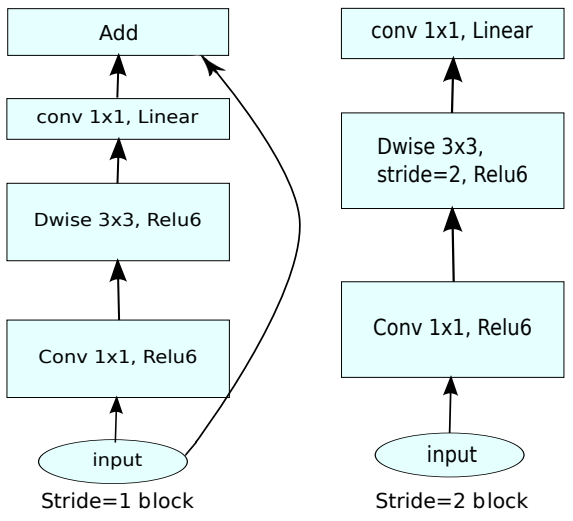}
    \caption{MobileNetV2 Convolutional Block \cite{RN33}}
    \label{fig:fig_d}
\end{figure}

MobileNetV2 \cite{RN33} is a general-purpose neural network built upon the ideas of depth-wise separable convolution in MobileNetV1 \cite{RN34}. However, MobileNetV2 introduces some new features: the inverted residual with linear bottleneck, and shortcut connections between the bottlenecks, which is used in many computer vision tasks such as classification, object detection, and semantic segmentation. The convolutional block of this architecture is shown in Figure \ref{fig:fig_d}. We use MobileNetV2 as the backbone of our model to extract intermediate- and high-level feature maps from the input scene. We initialize the weights of this neural network using the wights that are trained on ImageNet \cite{RN1}.\\

\subsection{U-Net Architecture}
This U-Net architecture was proposed by Ronneberger et al. \cite{RN3} to overcome the hardship of training deep models on very few training images and to enable more precise localization in the image segmentation task. It was developed for biomedical image segmentation. This architecture was built upon an FCN \cite{RN35}. The U-Net architecture is composed of two paths that look like a U character: contraction path and expansion path. The contraction path is just a stack of convolutional and max-pooling layers and acts as an encoder. This encoder is used to extract features and to capture the contextual information of the input scene. Expansion path acts as a decoder which is used to enable precise localization using upsampling operators and transposed convolutions. It increases the resolution of the output layer that improves the model performance by compensating for the lower spatial resolution of later layers in deep CNNs. It also enables the model to propagate the contextual information to the higher resolution layers and fuse the contextual information and local features maps. This structure also enables us to use the transfer learning for our backend model. For saliency prediction task, there is not enough training data for training a deep model. Transfer learning somewhat compensates for the lack of enough training data.\\
We introduce some changes in the U-Net architecture to make it appropriate for fast saliency prediction. The original U-Net architecture uses a copy operator to collect feature maps from each resolution level. The feature maps in some of these resolution levels have a large number of channels which is not desirable, because it increases the number of model parameters. Instead of the copy operator, we use convolutional layers with limited number of kernels to decrease the number of feature channels and to keep the model as compact as possible. Our experimental tests show that this change was very effective in reducing the number of model parameters and as a result reduce the required FLOPs. \\

\subsection{CDC layers}
Central difference convolutional network (CDCN) was introduced by Yu et al. \cite{RN4} for extracting intrinsic spooﬁng patterns via aggregating both intensity and gradient information. It is believed that CDCN is able to provide a more robust modeling capacity \cite{RN4}. Central difference convolution can be calculated as:\\
\begin{equation}
y\left(p_{0}\right)=\sum_{p_{n} \in \mathcal{R}} w\left(p_{n}\right) \cdot\left(x\left(p_{0}+p_{n}\right)-x\left(p_{0}\right)\right)
\end{equation}\\
where $x$ and $y$ are input and output feature maps respectively. $p_0$ denotes current location in both input and output feature maps while $p_n$ enumerates the locations in $R$. $R$ in our implementation with 2$\times$3 kernel is f(1; 1); (1; 0); … ; (0; 1); (1; 1). To calculate both the intensity-level semantic information and gradient-level detailed \cite{RN4}:\\
\begin{equation}
y\left(p_{0}\right)=\theta \cdot \sum_{p_{n} \in \mathcal{R}} w\left(p_{n}\right) \cdot\left(x\left(p_{0}+p_{n}\right)-x\left(p_{0}\right)\right)
+(1-\theta) \cdot \sum_{p_{n} \in \mathcal{R}} w\left(p_{n}\right) \cdot x\left(p_{0}+p_{n}\right)
\end{equation}\\
where hyperparameter $\theta \epsilon$[0; 1] tradeoffs the contribution between intensity-level and gradient-level information. In each resolution-level, we employ a convolutional layer that uses central difference convolution to extract more consistent patterns and biologically motivated features and to capture some ﬁne-grained invariant information.\\

\subsection{2D fully-connected layer}
It has been observed that the human eye ﬁxations are strongly biased towards the center of the image \cite{RN5} that is often explained via photographer bias \cite{RN6} or an un-engaged observer \cite{RN7}. It has been shown that including cues like center bias often improves the performance of saliency models \cite{RN8}.
As aforementioned, the original U-Net architecture is an end-to-end FCN \cite{RN3}, i.e. it only contains convolutional layers \cite{RN35}. Convolutional layers are location-invariant (or shift-invariant) \cite{RN6}. Hence, FCNs are incapable of extracting location-dependent patterns and capturing the center bias of the eye fixations because of the global nature of this property \cite{RN6}. To account for center bias, \cite{RN9} and \cite{RN10} combined their predicted saliency map with a ﬁxed Gaussian blob. Kruthiventi et al. \cite{RN6} introduced an LBC filter for capturing location-dependent patterns. To address this shortcoming in our model, we modify the U-Net architecture to make it suitable for the saliency prediction task. We introduce a 2D fully-connected layer that consists of some fully connected neurons arranged in a 2D array. Each neuron in this layer is connected to all neurons in its previous convolutional layer. Because the neurons of this layer are not location-invariant (or shift-invariant), this layer is capable of capturing location-dependent patterns and the center bias.\\

\section{Experimental Setup}
In this section, some popular evaluation metrics, evaluation baselines, and saliency datasets are described and then implementation details are provided.

\subsection{Evaluation Metrics}
For measuring the saliency model performance, several measures have been introduced. They fall into two categories \cite{RN36}\cite{RN2}\cite{RN38}. Coefﬁcient (CC), Kullback-Leibler divergence (KL), Earth Mover’s Distance (EMD), Information Gain (IG), and Similarity or histogram intersection(SIM) are distribution-based. They statistically compare the fixation map and the predicted the saliency map. Normalized Scanpath Saliency (NSS), Area under ROC Curve (AUC), and its variants including AUC-Judd, AUC-Borji, and Shufﬂed AUC (sAUC) are location-based. 

\subsection{Datasets}
In this work, we train and evaluate our models over four datasets: the dataset of SALICON Challenge 2015, the dataset of SALICON Saliency Prediction Challenge (LSUN 2017), MIT300, and MIT1003. The dataset of SALICON Challenge 2015 and the dataset of SALICON Saliency Prediction Challenge (LSUN 2017) \cite{RN11} are based on the MS COCO image dataset \cite{RN12}. They consist of 10,000 images for training, 5,000 images for validation, and 5,000 images for test. We call these datasets SALICON 2015 and SALICON 2017, respectively. Presently the model evaluation over SALICON 2015 test set is not available because it has been closed by the provider.
The MIT300 \cite{RN13} dataset the MIT1003 \cite{RN14} dataset consist of 300 and 1003 color images of natural indoor and outdoor scenes respectively in JPG format. 

\subsection{Implementation Details}
For SALICON 2015 and SALICON 2017, we train our models on the training data and validate it on the validation set. For these datasets, a batch size of 15 samples is chosen for the training and validation phase. As suggested by the MIT Saliency Benchmark \cite{mitb}, for MIT300, we pre-train our models on the SALICON and then fine-tune it on MIT1003. For fine-tuning, we split MIT1003 randomly into 904 images for the training set and 99 images for the validation set and chose a batch size of 9 samples. For the pre-training and fine-tuning stages, we initialize the learning rate to 10-4 and decrease it every two epochs by a factor of 10. Finally, the models with the best validation loss are chosen for evaluation on the test set.

\section{Experimental Evaluation}
In this section we perform some experimental analyses to validate our model architecture and its components. We then compare our model with other state-of-the-art saliency models. This comparison is conducted in two phases. In the first phase we conduct a fair comparison. To address this, we re-implement some state-of-the-art saliency models under assumptions of the same input size, no bias component, and the same loss function. This comparison helps us compare the saliency models based on their structural capabilities. In the second phase, we compare our model with the original form of state-of-the-art saliency models over some popular saliency benchmarks.

\subsection{Model Ablation Analysis}
Here we evaluate the contribution of our model architecture and it components. We used the following loss function for training and validation phases: 

\begin{equation}
L\left(\tilde{y}, y^{d e n}\right)=KL\left(\tilde{y}, y^{den}\right)
\end{equation}

\noindent where $\widetilde{y}$ and $y^{den}$ are the predicted saliency map and the ground-truth density distribution, respectively. KL is the Kullback-Leibler divergence which is among the most popular saliency measures. Kullback-Leibler divergence (KL-D) can be used to calculate the difference between two probability distributions. If we interpret the predicted map P and ground-truth map G, it can be  computed as \cite{RN2}:

\begin{equation}
KL(P, G)=\sum_{i} G_{i} \log \left(\varepsilon+\frac{G_{i}}{\varepsilon+P_{i}}\right)
\end{equation}

where $\epsilon$ is a constant that is used for regularization and $i$ indexes the $i^{th}$ pixel. As can be seen, the KL score is asymmetric. A larger KL value shows a larger difference between predicted saliency map and fixation map while a KL score of zero indicates that the model is predicting the saliency values perfectly.

As aforementioned, we use the U-Net architecture for our model. With all the advantages that this architecture has, it also has some properties that are not desirable for developing a compact model. For instance, because of the use of copy operators to collect feature maps from each resolution level, the number channels of collected feature maps are very large and as a result the number of model parameters will be very high. Instead of using copy operators, we use convolutional layers with a limited number of kernels to decrease the number of feature channels and as a result the number of model parameters to keep the model as compact as possible. Table \ref{tab:my_tabel0} compares the number of parameters in our model for using the original U-Net architecture and the proposed modified version of it.

\begin{table}[ht]
    \centering
    \caption{The comparison of using the original and modified U-Net architect}
    \label{tab:my_tabel0}
    \begin{tabular}{lcccc}
        \hline
            	ENCODER	& U-Net	ARCHITECTURE	& \# PARAMETERS (M) & FLOPs (G)\\
        \hline
        MobileNetV2	& original	& 6.57 & 2.56 \\
        MobileNetV2	& modified	& 1.94 & 0.87 \\
        \hline
        \end{tabular}{}
\end{table}
As Table \ref{tab:my_tabel0} shows, our proposed modification of the U-Net architecture drastically reduces the number of model parameters and required FLOPs. Experimental tests show that proposed modification of the U-Net architecture slightly hinders the performance of the model, but the decrease in the number of parameters and FLOPs are worth this reduction in performance. We implemented our model with several configurations to validate the contribution of using the CDC and fully-connected layers. Table \ref{tab:my_tabel1} shows the setups of these configurations. All of these models benefit from our modified U-Net architecture. Model 2 also employs biologically-motivated features. The  CDC operator is biologically-motivated and acts like the center-surround structure (ie fovea) in the retina. Model 3 employs an additional component (fully-connected layer) to address the location-dependent features such as prior bias. We evaluated these models on SALICON 2017 and MIT1003 datasets. The evaluation result is presented in Table \ref{tab:my_tabel2}.

\begin{table}[ht]
    \centering
    \caption{The configuration of our models}
    \label{tab:my_tabel1}
    \begin{tabular}{lcccc}
        \hline
        	& ENCODER	& U-Net	ARCHITECTURE	& CDC LAYERS	& 2D FULLY-CONNECTED LAYER\\
        \hline
        OUR MODEL 1	& MobileNetV2	&  \cmark	& \xmark	& \xmark\\
        OUR MODEL 2	& MobileNetV2	&  \cmark	&  \cmark	& \xmark\\
        OUR MODEL 3	& MobileNetV2	&  \cmark	&  \cmark	&  \cmark\\
        \hline
        \end{tabular}{}
\end{table}

\begin{table}[ht]
    \centering
    \caption{The number of parameter and validation loss of our models}
    \label{tab:my_tabel2}
    \begin{tabular}{lcccc}
        \hline
         & \#PARAMETERS (M) & FLOPS (G) & SALICON 2017 - VAL LOSS & MIT1003 - VAL LOSS\\
        \hline
        OUR MODEL 1 & 1.94 & 0.873 & 0.2414 & 0.33261\\
        OUR MODEL 2 & 1.94 & 0.890 & 0.24113 & 0.33118\\
        OUR MODEL 3 & 2.1 & 0.905 & 0.23944 & 0.32505\\
        \hline
        \end{tabular}{}
\end{table}

Table \ref{tab:my_tabel2} suggests that by the inclusion of the CDC features in the saliency prediction task, the performance of the saliency model is improved. It also proves the benefit of using our 2D fully-connected in a fully-convolutional network such as U-Net architecture. Given that our main goal is to create a fast and compact model, just reviewing the performance of the model on saliency benchmarks is enough. Therefore, in addition to examining the performance of the model, we also consider the number of model parameters and required FLOPs for the new component. Table \ref{tab:my_tabel2} also shows that our introduced components do not increase the number of model parameters and required FLOPs much, and their slight increase can be ignored by considering the amount of performance improvement.

\subsection{Fair Comparison}

We compare the proposed model to the  state-of-the-art saliency models under a unified assumption in order to hold fair comparison of models. Toward this goal, we develop a comparison framework that makes a fair comparison between state of art saliency models possible. Our proposed framework address the various aspects of the saliency models and also provide a standard basis for fair saliency model comparison. In the following, we describe  this framework in detail. This framework is available publicly and can be used for future studies.

\subsubsection{Saliency Model Framework}
\label{sec:framework}

The proposed framework consists of four components: (1) model interface, (2) data set classes, (3) loss functions, and (4) train and inference scripts. The heart of the framework is a common interface that facilitates implementation of any saliency model. We implemented 5 top performing saliency models. Saliency models are selected and re-implemented within our framework. These models are the following:

\noindent\textbf{SALICON:} SALICON \cite{RN15} is the first successful end-to-end trainable deep saliency model. It consists of two-stream coarse and fine grade image input that go through VGG neural architecture \cite{simonyan2014very} to produce two feature maps. Then, the features are normalized to the same spatial size and fused via a linear integration to produce a saliency map.

\noindent\textbf{DeepFix:} DeepFix \cite{RN6} is based on VGG neural architecture with dilated convolutions followed by two inception modules. DeepFix considers only one resolution input and incorporates 16 location bias filters. The location bias filters learn the center-bias from the data. In our implementation, we are omitting the center-bias layers.

\noindent\textbf{DeepGaze II:} This model \cite{RN10} extracts features from an image using VGG-19 network at several layers and concatenates them, resulting in a feature map with 2560 channels. It combines the features through a readout architecture. The readout network consists of four layers of 1x1 convolutions. In contrast to other models, DeepGaze model does not fine-tune VGG-19 feature extraction. In its original form, it adds also a center-bias to the final predicted maps. In our experiments, we avoid adding center-bias to the maps. 

\noindent\textbf{ML-Net:} ML-Net \cite{RN16} has a similar to DeepGaze II feature extraction pipeline. Instead of the DeepGaze II readout network, it uses a 3x3 convolution layer followed by 1x1 convolution layer to combine features and predict saliency. It also fine-tunes the VGG features for saliency prediction. In our implementation, we avoid adding center-bias.

\noindent\textbf{SAMResNet:} Saliency Attentive Model \cite{RN17} is the most distinct model from the rest of the above because of having the most sophisticated mechanism for inferring saliency. This model, akin to the rest, first extracts features using a modified ResNet \cite{He_2016_CVPR} architecture. This architecture exploits dilated convolutions. It, then, employs an attentive LSTM architecture to learn a temporally weighted feature map that is fused into a saliency map in combination with center-bias priors. In our implementations, we do not include the center-prior learning mechanism for the sake of comparison. 

\noindent\textbf{Loss Function and Training Procedure for Fair Comparison}:
Training procedure and loss functions are yet another important factor that influence saliency model performance. Thus, we employ Kullback–Leibler (KL) divergence as loss function for all models in the fair comparison phase. We also apply the same training procedure and starting from the same random seed initialization for all the models.

\noindent\textbf{Center-bias Free Comparison:} Center-bias\cite{RN5} is known as one factor influencing performance of saliency models. Many models are exploiting this properties in order to boost their performance with regard to prediction correctness. Some deep learning models are moving further and incorporate mechanisms for learning such bias from data. We, however, implement all the models without center-bias and avoid exploiting center-bias in our framework. Thus, our comparisons is fair with respect to center-bias exploit.

\noindent\textbf{Model Input Size:} The input size influences the number of multiplications in computing saliency maps. We, thus, report the FLOPS for equal input size image for all models. That is, we normalize the computation per pixel with respect to images with $240\times240$ dimension.

\subsubsection{Fair Comparison Results}
As aforementioned, to conduct a fair comparison, we use a framework that train, evaluate, and compares the saliency models under unified assumptions for the model interface, architecture, data set class, and training setup. By using this framework we fairly compare our model performance with some prominent saliency models. In table \ref{tab:my_tabel2}, three version for our proposed architecture is presented. Given that the 2D fully-connected layer is location-dependent and is capable of capturing bias information, for fair comparison phase we use model 2 to abide by the fair comparison assumptions. Model 2 has key properties of our proposed architecture, but do not has the bias capturing 2D fully-connected layer. By means of our framework, we compare the performance of our proposed architecture with the architecture of some prominent saliency models such as SALICON \cite{RN15},  DeepFix \cite{RN6}, DeepGaze II \cite{RN10}, ML-Net \cite{RN16}, SAMResNet \cite{RN17} under some unified assumptions. Tables \ref{tab:my_tabel3} and \ref{tab:my_tabel4}.

\begin{table}[ht]
    \centering
    \caption{Fair comparison of saliency models over SALICON 2015 validation set}
    \label{tab:my_tabel3}
    \begin{tabular}{lcccc}
        \hline
         & NSS & CC & AUC & SAUC\\
        \hline
        \textbf{OUR MODEL 2} & 2.801 & 0.805 & 0.883 & 0.777\\
        SALICON \cite{RN15} & 2.69 & 0.794 & 0.8756 & 0.8034\\
        ML-NET \cite{RN16} & 2.67 & 0.784 & 0.874 & 0.7987\\
        SAM-RESNET \cite{RN17} & 2.56 & 0.799 & 0.8771 & 0.8131\\
        DEEPFIX \cite{RN6} & 2.39 & 0.748 & 0.8696 & 0.7993\\
        DEEPGAZE II \cite{RN10} & 1.79 & 0.612 & 0.8538 & 0.8189\\
        \hline
        \end{tabular}{}
\end{table}

\begin{figure}[htb]
    \centering
    \includegraphics[width=0.4\textwidth]{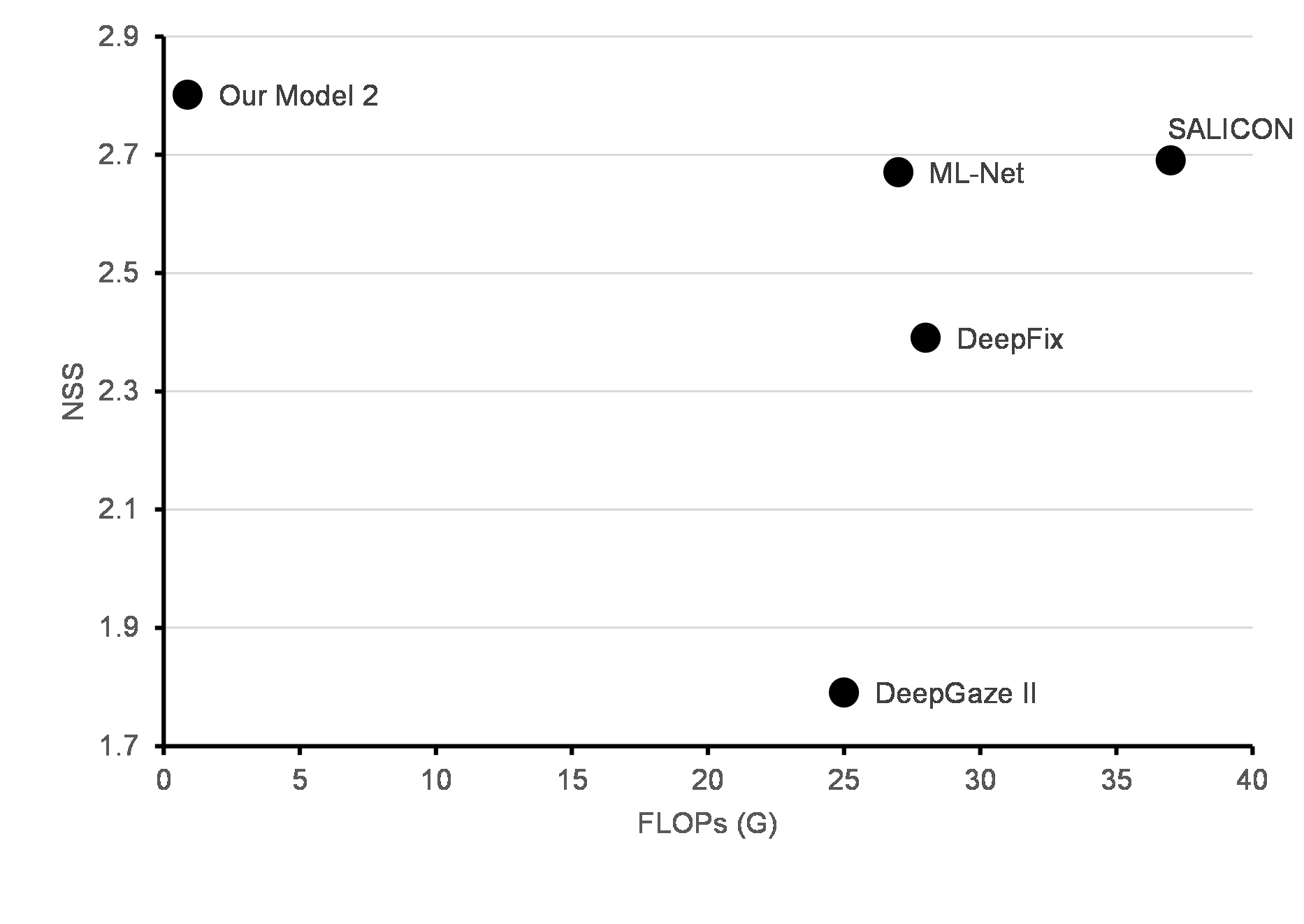}
    \includegraphics[width=0.4\textwidth]{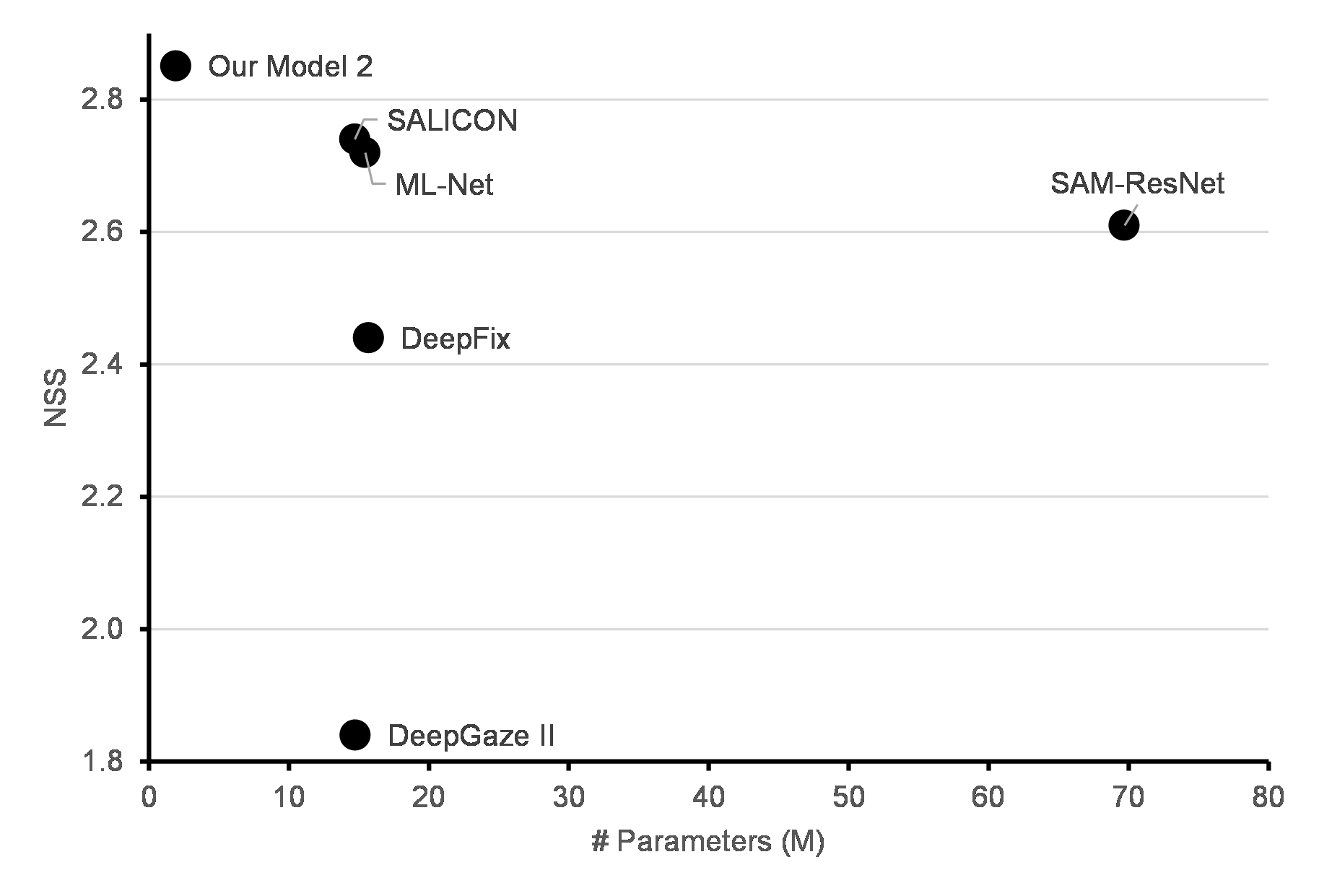}
    \caption{Left) NSS vs FLOPs, Right) NSS vs the number of parameters}
    \label{fig:fig_e}
\end{figure}

\begin{figure}[htb]
    \centering
    \includegraphics[width=0.4\textwidth]{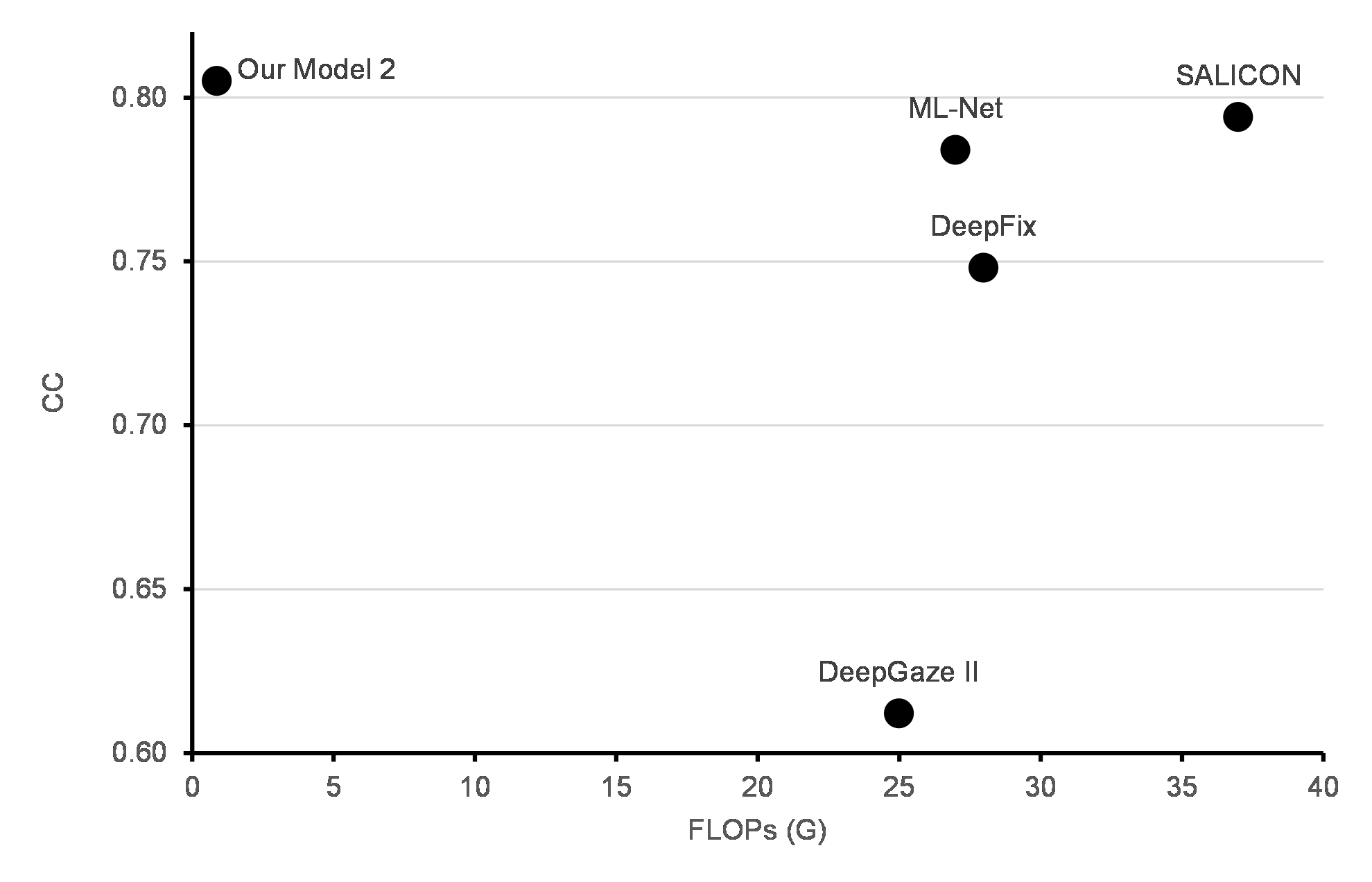}
    \includegraphics[width=0.4\textwidth]{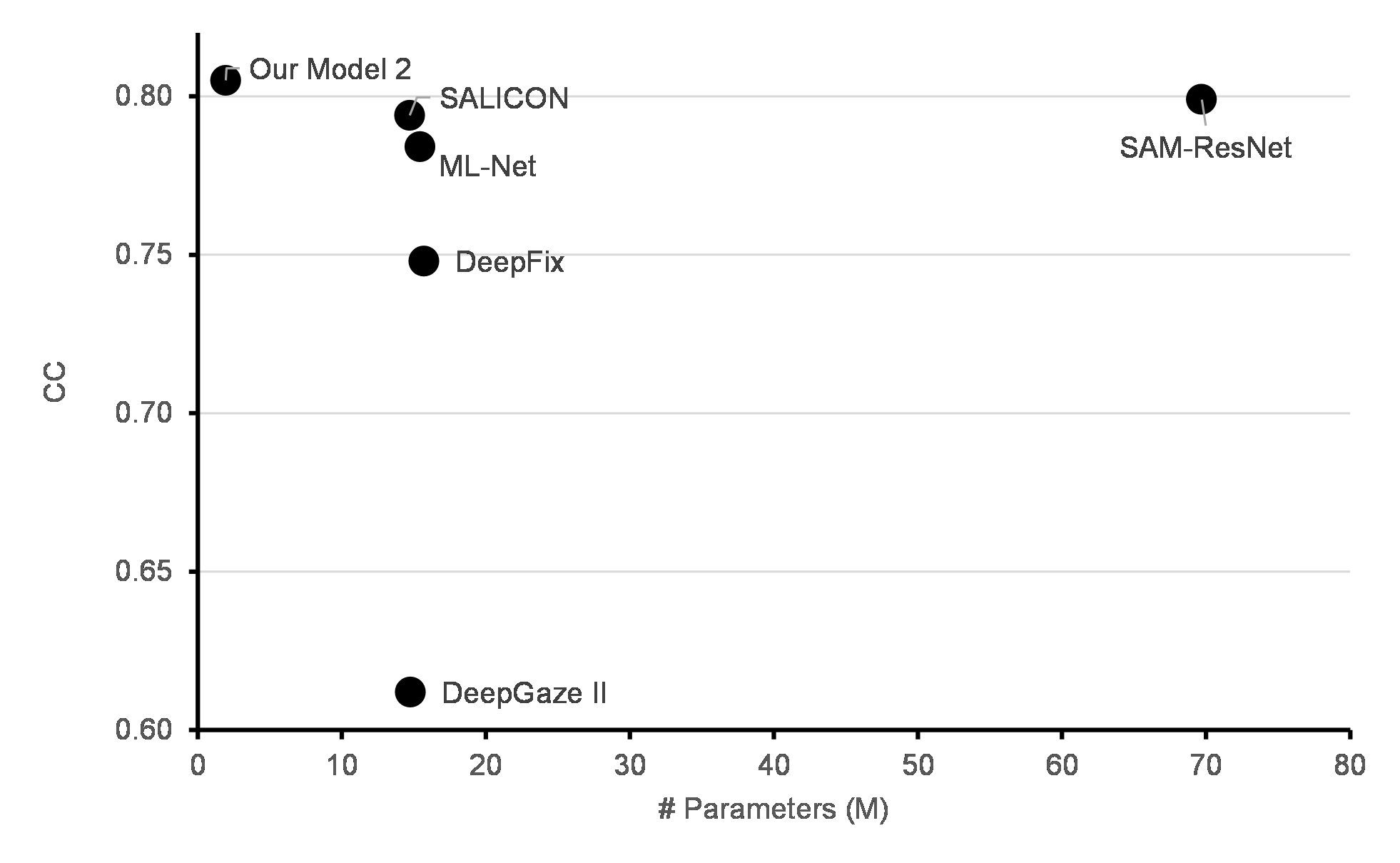}
    \caption{Left) CC vs FLOPs, Right) CC vs the number of parameters}
    \label{fig:fig_g}
\end{figure}

\begin{figure}[htb]
    \centering
    \includegraphics[width=0.4\textwidth]{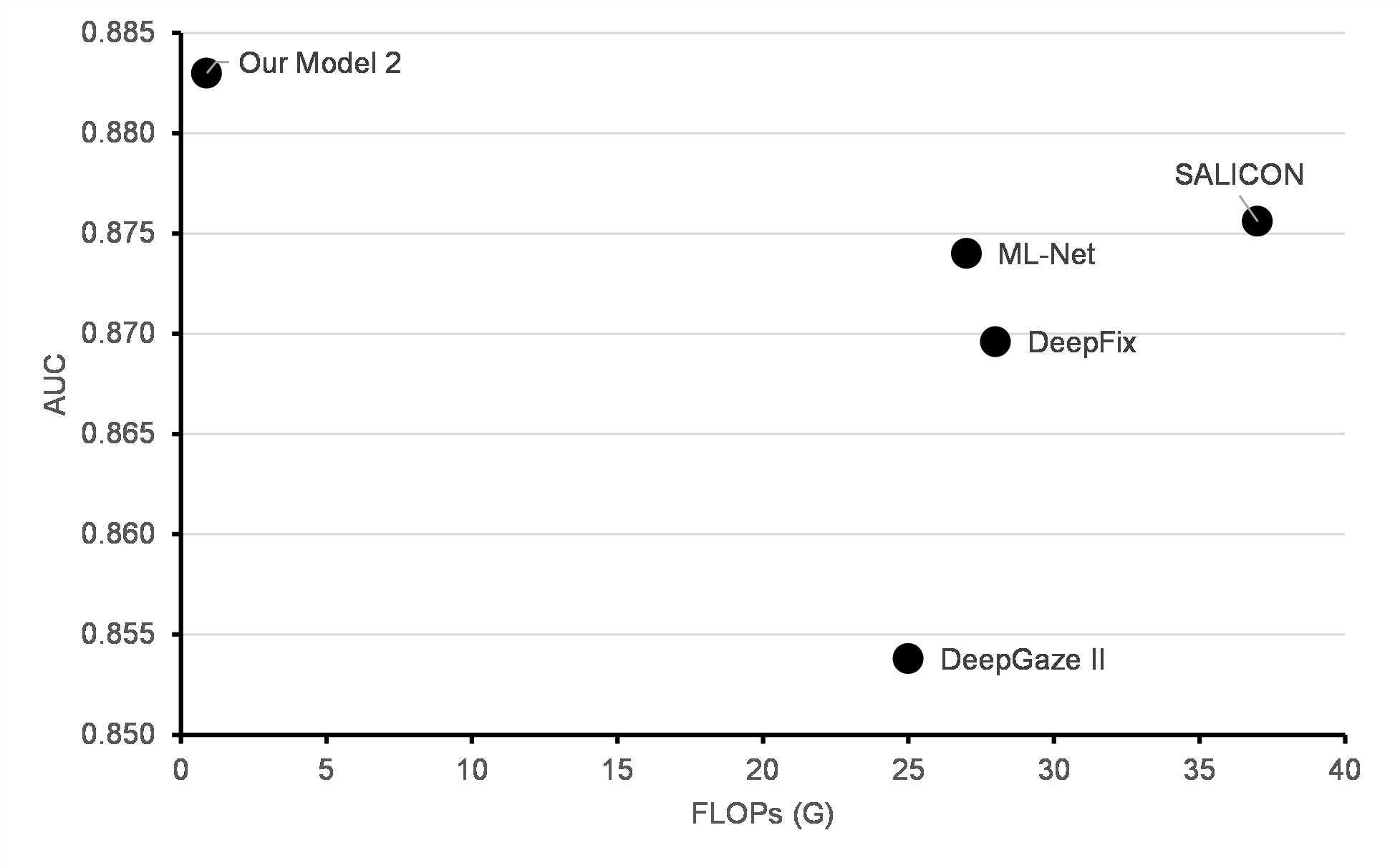}
    \includegraphics[width=0.4\textwidth]{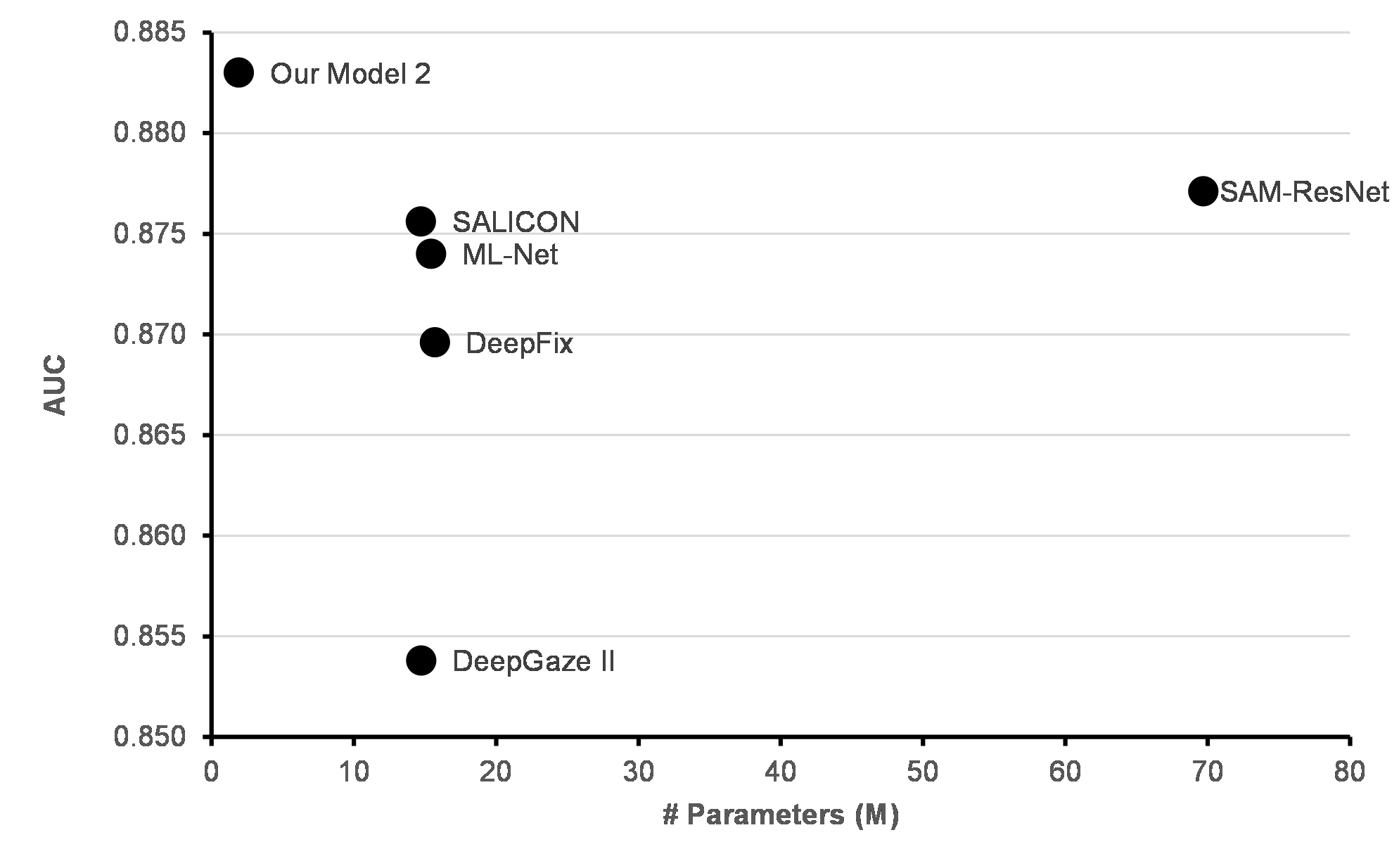}
    \caption{Left) AUC vs FLOPs, Right) AUC vs the number of parameters}
    \label{fig:fig_i}
\end{figure}

Table \ref{tab:my_tabel4} compares the number of parameters and FLOPs in our models with some state-of-the-art models. Accordingly, our models have far fewer parameters and FLOPs than these state-of-the-art saliency models. This property makes our models suitable for real-time application especially for the platforms with poor hardware performance such as mobile phones. Figures \ref{fig:fig_e}-\ref{fig:fig_i} present the performance of the models under the fair comparison assumption according to the number of parameters and FLOPs over SALICON 2015 .\\
\begin{table}[htb]
    \centering
    \caption{ The number of parameters and FLOPs of saliency modelst}
    \label{tab:my_tabel4}
    \begin{tabular}{lcc}
        \hline
         & \#PARAMETERS (M) & FLOPS (G)\\
        \hline
        SAM-RESNET \cite{RN17} & 70 & -\\
        SAM-VGG \cite{RN17} & 51.8 & -\\
        DEEPFIX \cite{RN6} & 15.7 & 28\\
        ML-NET \cite{RN16} & 15.45 & 27\\
        DEEPGAZE II \cite{RN10} & 14.76 & 25\\
        SALICON \cite{RN15} & 14.72 & 37\\
        \textbf{OUR MODEL 2} & 1.94 & 0.89\\
        \hline
        \end{tabular}{}
\end{table}

\subsection{Comparison with original models}
In this section, we compare the performance of our model with the original implementation of some state-of-the-art saliency prediction models. We used the third version of our architecture (model 3) as the final version of our proposed model. As presented in \ref{tab:my_tabel2}, our model 3 has a U-Net architecture with MobileNetV2 as the backend. It also uses CDC layers and fully-connected layers to extract robust and location-dependent features from the scene.
For this phase we used the following loss function to train and to evaluate our model:

\begin{equation}
L\left(\tilde{y}, y^{den}\right)=\alpha C C\left(\tilde{y}, y^{den}\right)+\beta K L\left(\tilde{y}, y^{den}\right)
\end{equation}

\noindent $\widetilde{y}$
where $\widetilde{y}$ and $y^{den}$ are the predicted saliency map and the ground-truth density distribution, respectively. CC and KL are the normalized scanpath saliency, the linear correlation coefficient, and the Kullback-Leibler divergence respectively which are among the most popular saliency measures. Loss parameters a and b have been set to -1, and 5, respectively. The number of parameters for the original implementation of  state-of-the-art saliency models has been calculated as Table \ref{tab:my_tabel5}. To conduct this comparison fair, we set the input size of all models to 240x240. 
\begin{table}[htb]
    \centering
    \caption{ The number of parameters of saliency models}
    \label{tab:my_tabel5}
    \begin{tabular}{lc}
        \hline
          & \#Parameters (M)\\
        \hline
        SAM-ResNet & 70\\
        SAM-VGG & 51.8\\
        salNet & 20.7\\
        ML-Net & 15.5\\
        EML-NET & 23.5\\
        \textbf{Our Model} & 2.1\\
        \hline
        \end{tabular}{}
\end{table}

\begin{figure}[htb]
    \centering
    \includegraphics[width=0.4\textwidth]{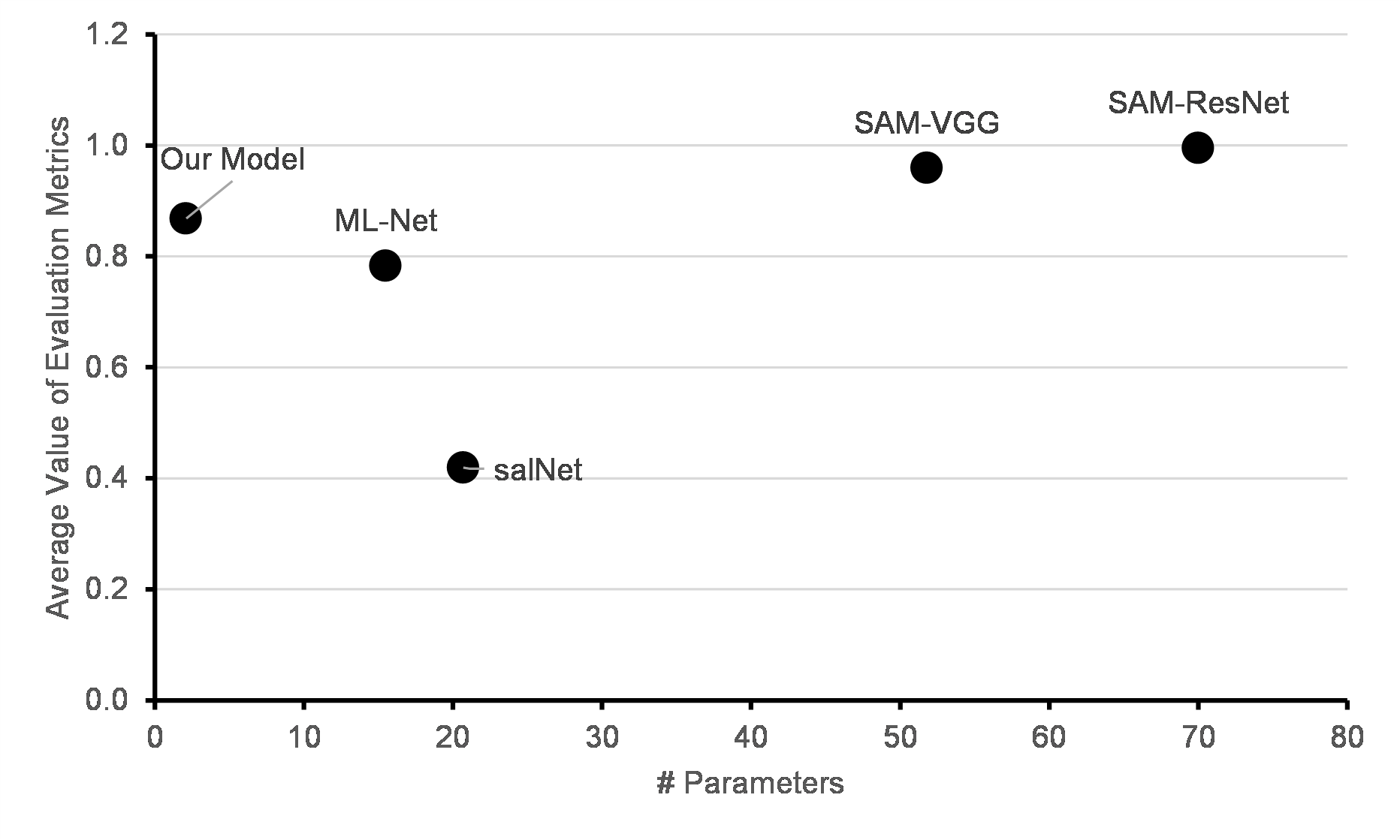}
    \includegraphics[width=0.4\textwidth]{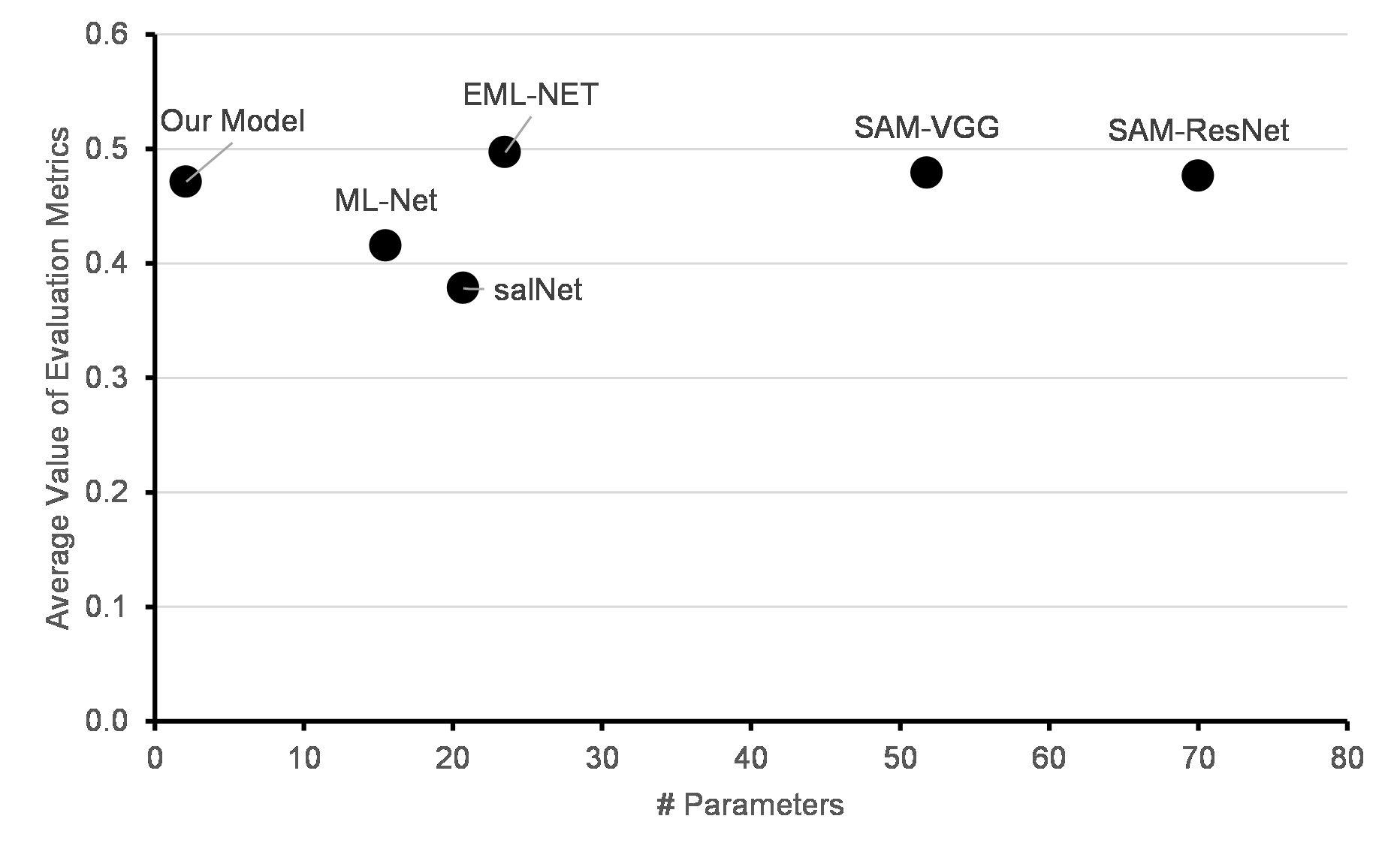}
    \caption{Left) Average value of evaluation metrics on SALICON 2015 validation set vs. the number of model parameters, Right) Average value of evaluation metrics on MIT300 vs the number of model parameters.}
    \label{fig:fig_k}
\end{figure}

The evaluation result of our model compared to some state-of-the-art saliency models over the SALICON 2015 validation set is reported in Table \ref{tab:my_tabel6}.To make the comparison over SALICON 2015 validation set easier, we calculated the average value of all evaluation metrics which is reported in the final column of Table \ref{tab:my_tabel6}. To calculate this average value, we normalized the value of each evaluation metric between 0 and 1. Then, we calculated the average value of all normalized metrics. This evaluation result is presented in Figure \ref{fig:fig_k}. The evaluation result of our model compared to some state-of-the-art saliency models over the SALICON 2017 test set is reported in Table \ref{tab:my_tabel7}.

\begin{table}[htb]
    \centering
    \caption{ Performance of saliency models over the SALICON 2015 validation set, compiled from SALICON Challenge 2015 website. The average is obtained over normalized values of metrics.}
    \label{tab:my_tabel6}
    \begin{tabular}{lccccc}
        \hline
         &	AUC	 & CC & SAUC & NSS & Average\\
         \hline
        SAM-ResNet \cite{RN17} & 0.886 & 0.844 & 0.787 & 3.26 & 0.995\\
        DSCLSTM \cite{RN18} & 0.887 & 0.835 & 0.788 & 3.221 & 0.988\\
        DSCLRCN \cite{RN18} & 0.887 & 0.835 & 0.785 & 3.221 & 0.983\\
        SAM-VGG \cite{RN17} & 0.883 & 0.83 & 0.782 & 3.219 & 0.959\\
        \textbf{Our Model} & 0.884 & 0.803 & 0.775 & 2.756 & 0.868\\
        ML-Net \cite{RN16} & 0.869 & 0.744 & 0.776 & 2.829 & 0.783\\
        DeepGaze II \cite{RN10} & 0.886 & 0.505 & 0.767 & 1.34 & 0.456\\
        SalNet: Deep convnet \cite{RN20} & 0.858 & 0.609 & 0.727 & 1.822 & 0.419\\
        SalNet: Shallow convnet \cite{RN20} & 0.817 & 0.548 & 0.658 & 1.625 & 0.069\\

        \hline
        \end{tabular}{}
\end{table}

The evaluation result of our model compared to some state-of-the-art saliency models over the MIT300 is reported in Table \ref{tab:my_tabel8}. To make the comparison over MIT300 easier, we calculated the average value of all evaluation metrics which is reported in the final column of Table \ref{tab:my_tabel8}. To calculate this average value, we normalized the value of each evaluation metric between 0 and 1. The normalized value of KL and EMD are multiplied by a minus to reverse their nature. Then, we calculated the average value of all normalized metrics. This evaluation result is presented in Figure \ref{fig:fig_k}. Overall, the results indicate superior performance considering both computational cost and prediction correctness.

\begin{table}[htb]
    \centering
    \caption{ Performance of saliency models over the SALICON 2017 test set}
    \label{tab:my_tabel7}
    \begin{tabular}{lccccccc}
        \hline
          & KLdiv & CC & AUC & NSS & SIM & IG & sAUC\\
        \hline
        MSI-Net \cite{RN21} & 0.307 & 0.889 & 0.865 & 1.931 & 0.784 & 0.793 & 0.736\\
        \textbf{Our Model} & 0.365 & 0.875 & 0.862 & 1.863 & 0.772 & 0.716 & 0.732\\
        GazeGAN \cite{RN22} & 0.376 & 0.879 & 0.864 & 1.899 & 0.773 & 0.72 & 0.736\\
        EML-NET \cite{RN23} & 0.52 & 0.886 & 0.866 & 2.05 & 0.78 & 0.736 & 0.746\\
        MD-SEM \cite{RN24} & 0.568 & 0.868 & - & 2.058 & - & - & -\\
        SAM-Resnet \cite{RN17} & 0.61 & 0.899 & 0.865 & 1.99 & 0.793 & 0.538 & 0.741\\
        SalNet \cite{RN20} & - & 0.622 & - & 1.859 & - & - & -\\
        \hline
        \end{tabular}{}
\end{table}

\begin{table}[htb]
    \centering
    \caption{Performance of our models compared to state-of-the-art saliency models over the MIT300 dataset, compiled from \cite{RN25}}
    \label{tab:my_tabel8}
    \begin{tabular}{lccccccccc}
        \hline
           & SIM & EMD & sAUC & CC & AUC-B & NSS & AUC-J & KL & Average\\
           \hline
            Baseline: infinite & 1 & 0 & 0.81 & 1 & 0.88 & 3.29 & 0.92 & 0 & 0.75\\
            DeepFix \cite{RN6} & 0.67 & 2.04 & 0.71 & 0.78 & 0.8 & 2.26 & 0.87 & 0.63 & 0.497\\
            EML-NET \cite{RN23} & 0.68 & 1.84 & 0.7 & 0.79 & 0.77 & 2.47 & 0.88 & 0.84 & 0.497\\
            DSCLRCN \cite{RN18} & 0.68 & 2.17 & 0.72 & 0.8 & 0.79 & 2.35 & 0.87 & 0.95 & 0.497\\
            SALICON \cite{RN15} & 0.6 & 2.62 & 0.74 & 0.74 & 0.85 & 2.12 & 0.87 & 0.54 & 0.495\\
            SAM-VGG \cite{RN17} & 0.67 & 2.14 & 0.71 & 0.77 & 0.78 & 2.3 & 0.87 & 1.13 & 0.479\\
            SAM-ResNet \cite{RN17} & 0.68 & 2.15 & 0.7 & 0.78 & 0.78 & 2.34 & 0.87 & 1.27 & 0.476\\
            \textbf{Our Model} & 0.65 & 2.32 & 0.71 & 0.74 & 0.8 & 2.1 & 0.86 & 0.8 & 0.471\\
            SalGAN \cite{RN27} & 0.63 & 2.29 & 0.72 & 0.73 & 0.81 & 2.04 & 0.86 & 1.07 & 0.467\\
            PDP \cite{RN8} & 0.6 & 2.58 & 0.73 & 0.7 & 0.8 & 2.05 & 0.85 & 0.92 & 0.454\\
            ML-Net \cite{RN16} & 0.59 & 2.63 & 0.7 & 0.67 & 0.75 & 2.05 & 0.85 & 1.1 & 0.416\\
            SalNet \cite{RN20} & 0.52 & 3.31 & 0.69 & 0.58 & 0.82 & 1.51 & 0.83 & 0.81 & 0.378\\
            Deep Gaze 2 \cite{RN10} & 0.46 & 3.98 & 0.72 & 0.52 & 0.86 & 1.29 & 0.88 & 0.96 & 0.377\\
            GBVS \cite{RN25} & 0.48 & 3.51 & 0.63 & 0.48 & 0.8 & 1.24 & 0.81 & 0.87 & 0.308\\
            Deep Gaze 1 \cite{RN9} & 0.39 & 4.97 & 0.66 & 0.48 & 0.83 & 1.22 & 0.84 & 1.23 & 0.288\\
            eDN \cite{RN28} & 0.41 & 4.56 & 0.62 & 0.45 & 0.81 & 1.14 & 0.82 & 1.14 & 0.265\\
            IttiKoch2 \cite{RN29} & 0.44 & 4.26 & 0.63 & 0.37 & 0.74 & 0.97 & 0.75 & 1.03 & 0.222\\
            Baseline: Center & 0.45 & 3.72 & 0.51 & 0.38 & 0.77 & 0.92 & 0.78 & 1.24 & 0.199\\
            Baseline: 1 human & 0.38 & 3.48 & 0.63 & 0.52 & 0.66 & 1.65 & 0.8 & 6.19 & 0.157\\
            SUN saliency \cite{RN30} & 0.38 & 5.1 & 0.61 & 0.25 & 0.66 & 0.68 & 0.67 & 1.27 & 0.107\\
            IttiKoch \cite{RN31} & 0.2 & 5.17 & 0.53 & 0.14 & 0.54 & 0.43 & 0.6 & 2.3 & -0.059\\
            Baseline: Perm. & 0.34 & 4.59 & 0.5 & 0.2 & 0.59 & 0.49 & 0.68 & 6.12 & -0.065\\
            Baseline: Chance & 0.33 & 6.35 & 0.5 & 0 & 0.5 & 0 & 0.5 & 2.09 & -0.147\\
        \hline
        \end{tabular}{}
\end{table}

\begin{figure}[htb]
    \centering
    \includegraphics[width=0.5\textwidth]{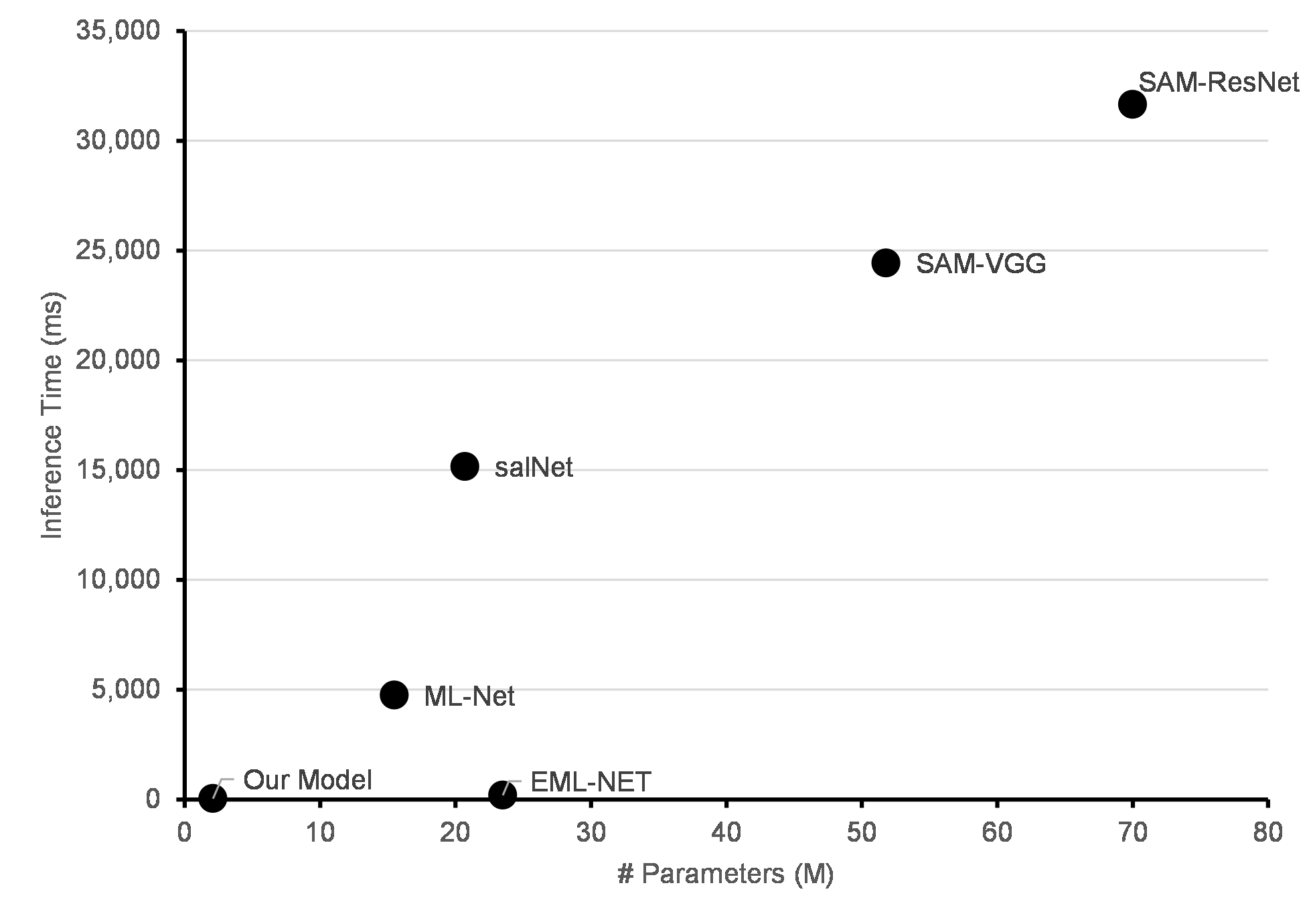}
    \caption{Inference time vs the number of model parameters}
    \label{fig:fig_m}
\end{figure}

Table \ref{tab:my_tabel9} shows the inference time of our models compared to the implementation of the original form of some state-of-the-art saliency prediction models on CPU Intel Xeon 2.30GHz. The Inference Time reports the run time for the model to predict the saliency map of an input image. These measurements are the average run time over several input images. To keep the comparison fair, we measure the inference time of the models under the assumption of the input size of 240x240. We also presented this evaluation result in Figure \ref{fig:fig_m}.

\begin{table}[htb]
    \centering
    \caption{Speed of our models in prediction phases using CPU compared to some other state-of-the-art models}
    \label{tab:my_tabel9}
    \begin{tabular}{lc}
        \hline
        &	Inference Time (ms)\\
        \hline
        SAM-ResNet \cite{RN17} & 31,660\\
        SAM-VGG \cite{RN17} & 24,426\\
        SalNet \cite{RN20} & 15,168\\
        DSCLRCN \cite{RN18} & 9,023\\
        GBVS \cite{RN32} & 6,101\\
        ML-Net \cite{RN16} & 4,747\\
        SalGAN \cite{RN27} & 4,367\\
        IttiKoch2 \cite{RN29} & 301\\
        Deep Gaze 2 \cite{RN10} & 270\\
        EML-NET \cite{RN23} & 193\\
        \textbf{Our Model} & 37\\
        \hline
        \end{tabular}{}
\end{table}

\begin{figure}[htb]
    \centering
    \includegraphics[width=0.8\textwidth]{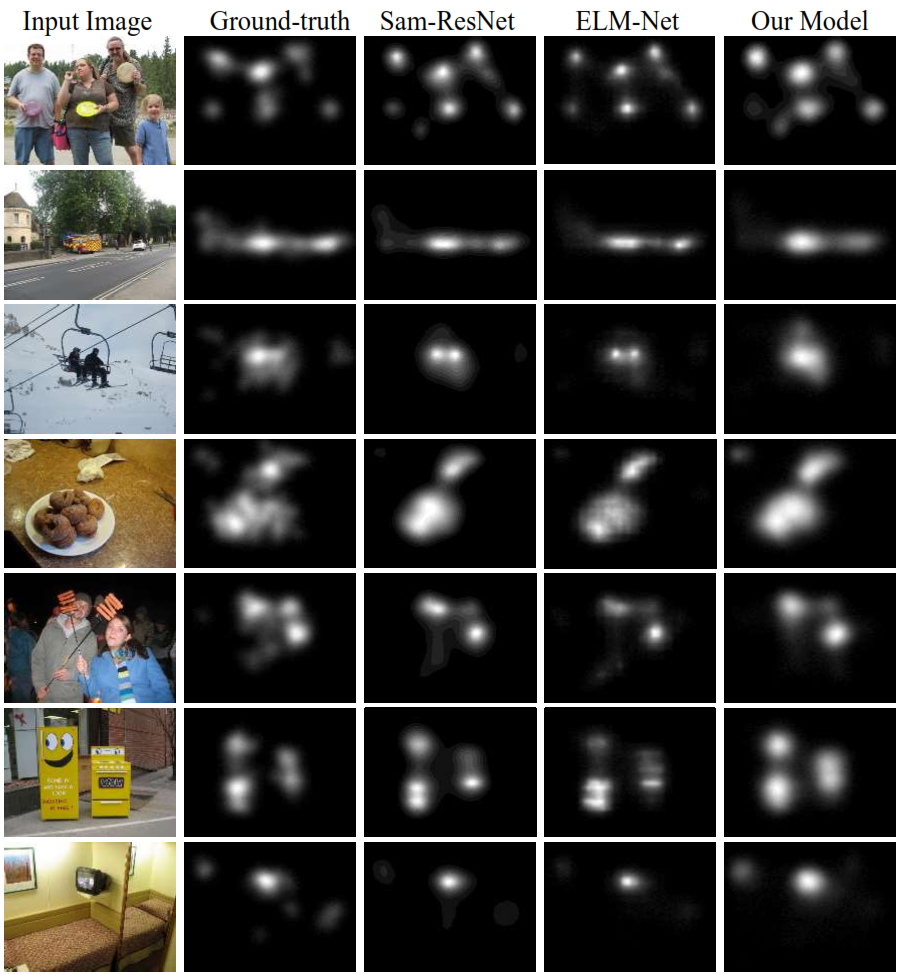}
    \caption{Qualitative results and comparison to the state of the art.}
    \label{fig:fig_sample}
\end{figure}

Table \ref{tab:my_tabel9} and Figure \ref{fig:fig_m} suggest that our models can predict the saliency maps hundreds of times faster than prominent state-of-the-art saliency models while maintaining competitive performance. Compared to SAM-ResNet and SAM-VGG models, our models have similar and in some cases, better saliency results while performing up to 1000x faster. In Figure \ref{fig:fig_sample}, we compare the output of our model with EML-NET \cite{RN23} and SAM-ResNet \cite{RN17}. These state-of-the-art saliency models have 23.5 and 70 million parameters respectively, while our model has only 2.1 million parameters. As indicated, our model is compact, requiring less memory footprint. It is fastest model, and preforms well in terms of salincy measures.

\section{Conclusion}
Most state-of-the-art saliency models suffer from a large number of parameters and are not suitable for real-time applications on CPU platforms. The evaluation results show the effect of our modification in U-net architecture for developing a compact deep saliency model. The original U-Net architecture uses a copy operator to collect feature maps from each resolution level. The feature maps in some of these resolution levels have a large number of channels which is not desirable because it increases the number of model parameters. To decrease the number of channels, we use convolutional layers with a limited number of kernels in our U-Net architecture. Convolutional layers use weight sharing and are shift-invariant. On the one hand, this weight sharing reduces the number of required parameters of the model. On the other hand, this property makes the model shift-invariant, that is, the fully convolutional models are incapable of capturing location-dependent information. To address this problem, we employed a 2D fully-connected layer to extract location-specific features. To extract biologically motivated and robust features, we employed CDC layers which leads to performance improvement. These components can be used for any other saliency model to include the aforementioned information.


Experimental results show that our model strikes a balance between correctness of prediction (saliency prediction goodness measures) and computational demand in terms of memory (number of parameters) and speeds (flops).  It only needs 0.9 Giga FLOPs, which is very suitable for real-time application especially for the platforms with constrained hardware specfication such as mobile phones and digital cameras. Table \ref{tab:my_tabel9} shows that our model can process 27 frames per second, which is suitable for many real-time tasks.

\bibliographystyle{unsrt}  



\end{document}